\documentclass[conference]{IEEEtran}
\IEEEoverridecommandlockouts
\usepackage{cite}
\usepackage{amsmath,amssymb,amsfonts}
\usepackage{algorithmic}
\usepackage{graphicx}
\usepackage{textcomp}
\usepackage[table]{xcolor}
\usepackage{xcolor}
\usepackage[pdftex]{hyperref}

\def\BibTeX{{\rm B\kern-.05em{\sc i\kern-.025em b}\kern-.08em
    T\kern-.1667em\lower.7ex\hbox{E}\kern-.125emX}}

\begin{document}

\title{Fractional norms and quasinorms do not help to overcome the curse of dimensionality
\thanks{The project is supported by the Ministry of Education and Science of the Russian Federation (Project No 14.Y26.31.0022).}
}

\author{\IEEEauthorblockN{1\textsuperscript{rd} Evgeny M. Mirkes}
\IEEEauthorblockA{\textit{Department of mathematics} \\
\textit{University of Leicester}\\
Leicester, UK \\
\textit{Lobachevsky State University} \\
Nizhni Novgorod, Russia \\
em322@leicester.ac.uk}
\and
\IEEEauthorblockN{2\textsuperscript{nd} Jeza Allohibi}
\IEEEauthorblockA{\textit{Department of mathematics} \\
\textit{University of Leicester}\\
Leicester, UK \\
\textit{Taibah University}\\
 Medina, Saudi Arabia\\
jhaa1@leicester.ac.uk }
\and
\IEEEauthorblockN{3\textsuperscript{st} Alexander N. Gorban}
\IEEEauthorblockA{\textit{Department of mathematics} \\
\textit{University of Leicester}\\
Leicester, UK \\
\textit{Lobachevsky State University} \\
Nizhni Novgorod, Russia \\
a.n.gorban@leicester.ac.uk}
}

\maketitle

\begin{abstract}
The curse of dimensionality causes the well-known and widely discussed  problems for machine learning methods. There is a hypothesis that using of the Manhattan distance and even fractional quasinorms lp (for p less than 1) can help to overcome the curse of dimensionality in classification problems. In this study, we systematically test this hypothesis. We confirm that fractional quasinorms have a greater relative contrast or coefficient of variation than the Euclidean norm l2, but we also demonstrate that the distance concentration shows qualitatively the same behaviour for all tested norms and quasinorms and the difference between them decays as dimension tends to infinity. Estimation of classification quality for kNN based on different norms and quasinorms shows that a greater relative contrast does not mean better classifier performance and the worst performance for different databases  was shown by different  norms (quasinorms). A systematic comparison shows that the difference of the performance of kNN based on lp for p=2, 1, and 0.5 is statistically insignificant.
\end{abstract}


\section{Introduction}
The term ``curse of dimensionality'' was introduced by Bellman \cite{bellman1961}. Nowadays, this is a general term for problems related to high dimensional data, for example, for Bayesian modelling \cite{bishop2006}, nearest neighbour prediction \cite{hastie2009} and search \cite{korn2001}, neural networks \cite{bishop1995}, and many others. Many authors \cite{beyer1999}, \cite{hinneburg2000}, \cite{aggarwal2001}, \cite{aggarwal2001outlier}, \cite{radovanovic2010} studied the ``meaningfulness'' of distance based classification in high dimensions. These studies are related to the distance concentration, which means that in high dimensional space the distances between almost all pairs of points have almost the same value.

The term ``blessing of dimensionality'' was introduced by Kainen in 1997 \cite{kainen1997utilizing}. The ``blessing of dimensionality'' considers the same distance concentration effect from the different point of view \cite{chen2013}, \cite{gorban2016}, \cite{gorban2018}, \cite{liu2017}, \cite{gorban2018correction}. The distance concentration was discovered in the foundation of statistical physics and analysed further in the context of probability theory, functional analysis, and  geometry (reviewed by \cite{GIANNOPOULOS2000}, \cite{gorban2018}, \cite{ledoux2001}, \cite{francois2007concentration}). The blessing of dimensionality allows us to use some specific high dimensional properties to solve problems \cite{donoho2000high},\cite{anderson2014more}. The important property is linear separability of points from  random sets in high dimensions \cite{gorban2018}, \cite{gorban2018correction}.

The  $l_p$ functional $\|x\|_p$   in $d$ dimensional  vector space is defined as
\begin{equation}
\|x\|_p=\left(\sum_{i=1}^{d}x_i^p\right)^{1/p} \label{eq:Norm}.
\end{equation}
It is a  norm for $p\ge1$ and a quasinorm for $0<p<1$ because of violation of the triangle inequality \cite{kothe1969topological}.
It is well known that for $p<q$ we have $\|x\|_p\ge \|x\|_q, \forall x$.

Measurement of dissimilarity and errors by subquadratic functionals reduces the  effect of outliers and can help to construct more robust data analysis methods \cite{franccois2005non}, \cite{aggarwal2001}, \cite{francois2007concentration}. Utilisation of these functionals for struggling with the curse of dimensionality was proposed in several works \cite{aggarwal2001}, \cite{francois2007concentration}, \cite{dik2014fractional}, \cite{jayaram2012can} \cite{france2012distance}, \cite{doherty2004non}.

In 2001, C.C. Aggarwal with co-authors \cite{aggarwal2001} described briefly the effect of using  fractional quasinorms for high-dimensional problems. They demonstrated that using of $l_p$ ($p\leq1$) can compensate the distance concentration. This idea was used further in many works \cite{cormode2002fast}, \cite{datar2004locality}, \cite{radovanovic2010}. One of the main problems of $l_p$ quasinorms usage for $p<1$ is time of calculation of minimal distances and solution of optimization problems with $l_p$ functional (which a even non-convex for $p<1$). Several methods have been developed to accelerate the calculations \cite{cormode2002fast}, \cite{gorban2018PQSQ}. The main outcome of \cite{aggarwal2001} was the use of Manhattan distance instead of Euclidean one \cite{Allen2014}, \cite{elkan2003using}, \cite{chang2003cbsa}. The main reason for this is the fact that for $p<1$ functional $l_p$ is not a norm but is a non-convex quasinorm. All methods and algorithms which assume triangle inequality \cite{elkan2003using}, \cite{demartines1994analyse}, \cite{yianilos1999excluded} cannot use such a quasinorm.

Comparison of different $l_p$ functionals for data mining problems is yet fragmental, see, for example, \cite{aggarwal2001}, \cite{singh2013k}, \cite{hu2016distance}. In our study, we performed systematic testing. In general case distance concentration for $l_p$ functionals was less for lower $p$ but for all $p$ the shape of distance concentration as a function of dimension is qualitatively the same. Moreover, the difference in distance concentration for different $p$ decreases with dimension increasing. We systematically tested the hypothesis that the measurement of dissimilarity by subquadratic norms $l_p (1\leq p<2)$ or even quasinorms $(0< p<1)$ can help to overcome the curse of dimensionality in classification problems. We demonstrated that these norms and quasinorms {\it do not} improve k Nearest Neighbour (kNN) classifiers in high dimensions systematically and significantly.

There are two main results in this study: (i) usage of $l_p$ functionals with small $p$ does not prevent the distance concentration and (ii) the smaller distance concentration does not mean the better accuracy of kNN classification.

The further part of our paper is organised as follows.
In Section `Measure concentration' we presented results of an empirical test of distance concentration for Relative Contrast (RC) and Coefficient of Variation (CV) also known as relative variance.
In Section `Dimension estimation' we presented description of 6 used dimensions.
In Section `Comparison of $l_p$ functionals' we describe the approaches used for $l_p$ functionals comparison, the used databases and the classification quality measures.
In Section `Dimension comparison' we presented results of 6 discussed dimensions comparison.
In Section `Results of $l_p$ functionals comparison' we presented results of the described tests to compare different $l_p$ functionals.
In Section `Discussion' discussion and outlook are presented.

All software and databases used for this study can be found in \cite{ourSoft}. Some results of this work were presented partially at IJCNN2019 \cite{mirkes2019fractional}.

\section{Measure concentration}\label{MesConc}

Let us consider a database $X$ with $n$ data points $X={x_1,\ldots,x_n}$ and $d$ real-valued attributes, $x_i=(x_{i1},\ldots,x_{id})$. We consider databases of two types: randomly generated database with i.i.d. components from the uniform distribution on the interval $[0,1]$ (this section) and real life databases  (Section \ref{compar}). The $l_p$ functional for vector $x$ is defined by \eqref{eq:Norm}. For comparability of results in this study we consider set of norms and quasinorms used in \cite{aggarwal2001} with one more quasinorm ($l_{0.01}$): $l_{0.01}, l_{0.1}, l_{0.5}, l_{1}, l_{2}, l_{4}, l_{10}, l_{\infty}$.

\begin{figure}[b]
\centerline{\includegraphics[width=0.5\columnwidth ]{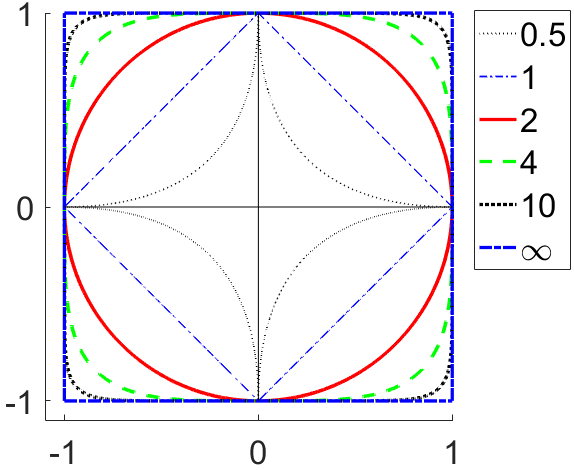}}
\caption{Unit level sets for $l_p$ functionals (``Unit spheres'').}
\label{fig:UnitCircles}
\end{figure}

Fig.~\ref{fig:UnitCircles} shows forms of unit level sets for all considered norms and quasinorms excluding $l_{0.01}$ and $l_{0.1}$. For these two quasinorms, graphs are visually indistinguishable from the central cross.

Several different indicators were used to study distance concentration:
\begin{itemize}
\item Relative Contrast (RC) \cite{beyer1999}, \cite{aggarwal2001}, \cite{francois2007concentration}
\begin{equation}
\text{RC}_p(X,y)=\frac{|\max_i \|x_i-y\|_p - \min_i \|x_i-y\|_p|}{\min_i \|x_i-y\|_p}; \label{eq:RC}
\end{equation}
\item Coefficient of Variations (CV) or relative variance \cite{demartines1994analyse}, \cite{yianilos1999excluded}, \cite{francois2007concentration}
\begin{equation}
\text{CV}_p(X,y)=\frac{\sqrt{var(\|x_i-y\|_p|)}}{mean(\|x_i-y\|_p)}, \label{eq:CV}
\end{equation}
where $var(x)$ is variance and $mean(x)$ is mean value of random variable $x$;
\item Hubness (popular nearest neighbours) \cite{radovanovic2010}.
\end{itemize}
In our study we use RC and CV.

Table 2  in \cite{aggarwal2001}  shows that fraction of cases where $\text{RC}_1>\text{RC}_2$ increases with dimension. It can be easily shown that for specific choice of $X$ and $y$ all three relations between RC$_1$ and RC$_2$ are possible $\text{RC}_1(X,y)>\text{RC}_2(X,y)$, $\text{RC}_1(X,y)=\text{RC}_2(X,y)$, or $\text{RC}_1(X,y)<\text{RC}_2(X,y)$. To evaluate the probabilities of these three outcomes, we performed the following experiment. We generated dataset $X$ with $k$ points and $100$ coordinates. Each coordinate of each point was uniformly randomly generated for the interval $[0, 1]$. For each dimension $d=1, 2, 3, 4, 10, 15, 20, 100$ we create $d$ dimensional database $X_d$ by selection of the first $d$ coordinates of points in $X$. We calculated $\text{RC}_p$ as the mean value of RC for each point of $X_d$:
\begin{equation*}
\text{RC}_p=\frac{1}{k}\sum_{i=1}^{k}\text{RC}_p(X_d\backslash \{x_i\},x_i),
\end{equation*}
where $X\backslash\{y\}$ is the database $X$ without point $y$. We repeated this procedure 1000 times and calculated the fraction of cases when $\text{RC}_1(X,y)>\text{RC}_2(X,y)$. Results of this experiment are presented in Table~\ref{tab:RCComp}. Table~\ref{tab:RCComp} shows that for $k=10$ points our results are very similar to the results  presented in  Table 2 in \cite{aggarwal2001}. Increasing of number of points shows that already for relatively small number of points ($k\approx 20$) for almost all databases $\text{RC}_1(X,y)>\text{RC}_2(X,y)$.

\begin{table}[tb]
\caption{Comparison of RC for $l_1$ and $l_2$ for different dimension (Dim) and different number of points}
\begin{center}
\begin{tabular}{|c|c|c|c|c|}
\hline
\textbf{Dim}&\multicolumn{4}{|c|}{\textbf{$P(\text{RC}_2<\text{RC}_1)$ for \# of points}} \\
\cline{2-5}
 & \textbf{10 \cite{aggarwal2001}} & \textbf{10} & \textbf{20} & \textbf{100} \\
\hline
  1 & 0 & 0 & 0 & 0 \\ \hline
  2 & 0.850 & 0.850 & 0.960 & 1.00 \\ \hline
  3 & 0.887 & 0.930 & 0.996 & 1.00 \\ \hline
  4 & 0.913 & 0.973 & 0.996 & 1.00 \\ \hline
 10 & 0.956 & 0.994 & 1.00 & 1.00 \\ \hline
 15 & 0.961 & 1.000 & 1.00 & 1.00 \\ \hline
 20 & 0.971 & 0.999 & 1.00 & 1.00 \\ \hline
100 & 0.982 & 1.000 & 1.00 & 1.00 \\ \hline

\end{tabular}
\label{tab:RCComp}
\end{center}
\end{table}

\begin{figure}[t]
\centerline{\includegraphics[width=0.9\columnwidth ]{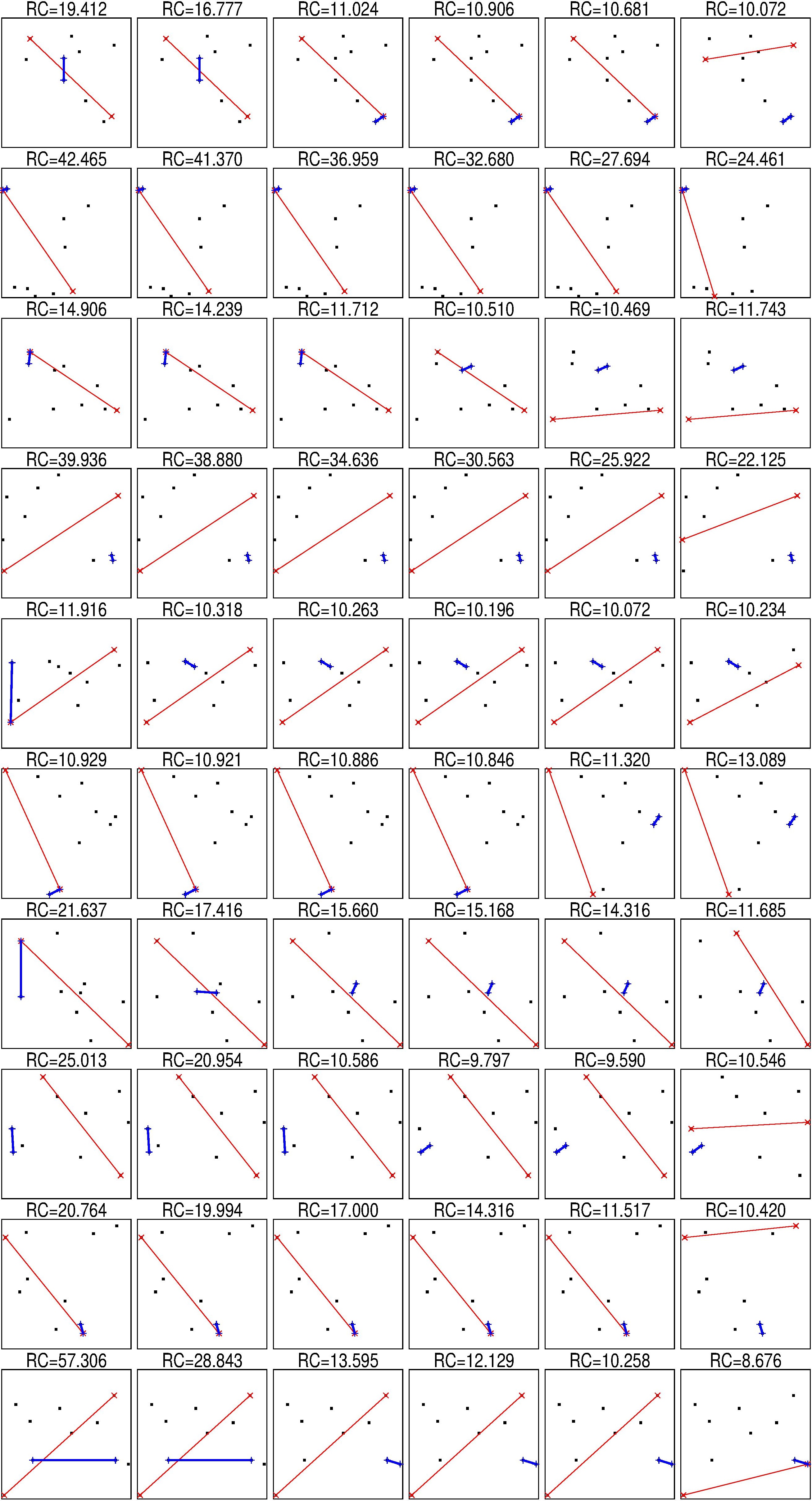}}\caption{10 randomly generated sets of 10 points, thin red line connects the furthest points and bold blue line connects closest points, columns (from left to right) corresponds to $p=0.01, 0.1, 0.5, 1, 2, \infty $ }
\label{fig:sets}
\end{figure}

This means that appearance of non-negligible fraction of  cases where $\text{RC}_2>\text{RC}_1$ is caused by very small size of a sample. For not so small samples we almost always have $\text{RC}_2<\text{RC}_1$. The main reason for this is different pairs of closest and furthest points for different metrics. Several examples of such sets are presented in Fig.~\ref{fig:sets}.
Fig.~\ref{fig:sets} shows that $\text{RC}_2<\text{RC}_\infty$ in rows 3, 5, 6, and 8 and $\text{RC}_1<\text{RC}_2$ in row 6. These results allows us to formulate a hypothesis that in general case almost always $\text{RC}_p<\text{RC}_q, \forall p>q$. RC widely used to study properties of finite set of points but for distributions of points the CV is more appropriate. We assume that for CV hypothesis $\text{CV}_p<\text{CV}_q, \forall p>q$ is also true.

\begin{figure}[t]
\centerline{\includegraphics[width=0.5\columnwidth ]{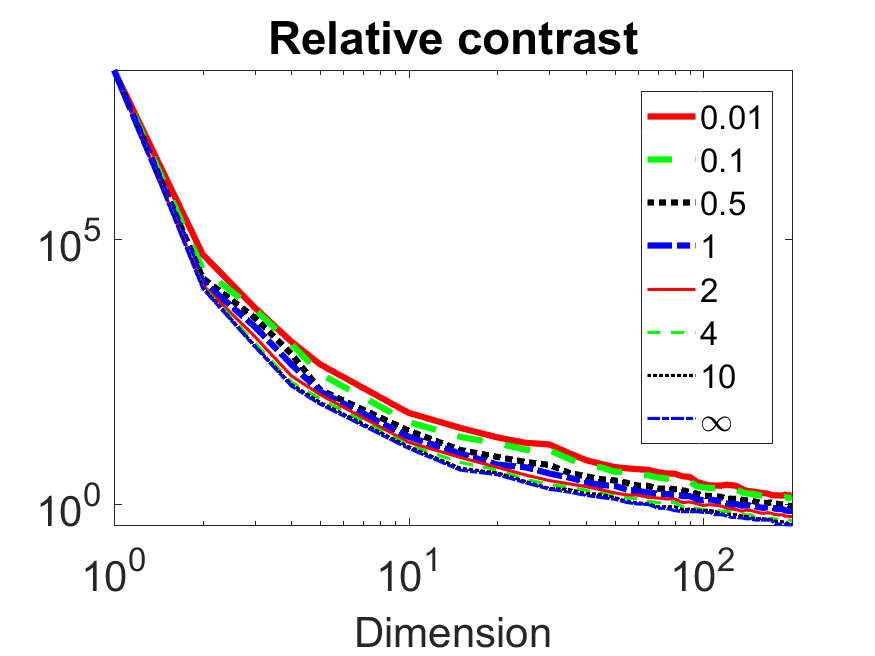}\includegraphics[width=0.5\columnwidth ]{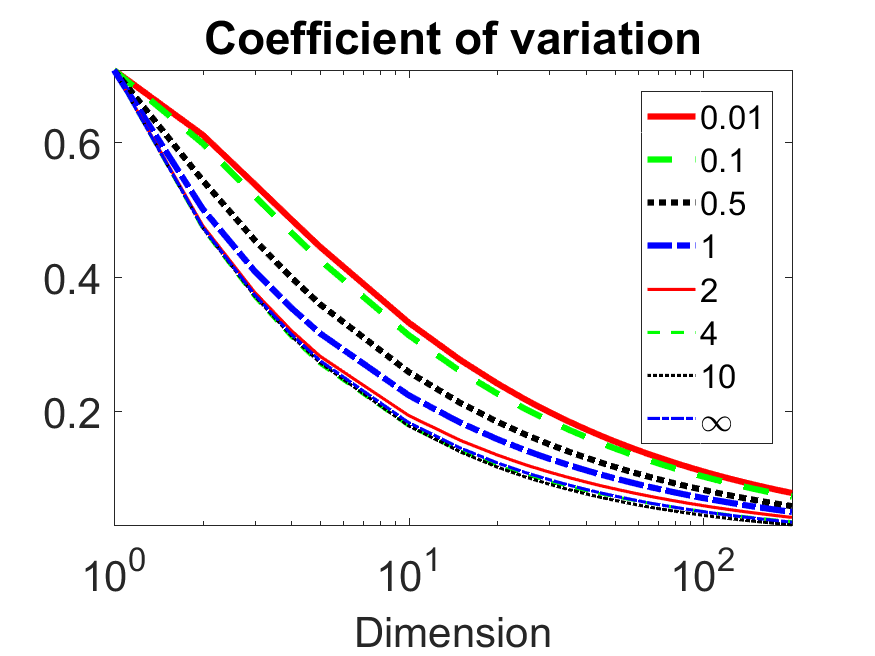}}
\caption{Changes of RC (left) and CV (right) with dimension increasing for several metrics}
\label{fig:RC1}
\end{figure}

To check this hypothesis we performed the following experiment. We generated database $X$ with 10,000 points in 200 dimensional space. Each coordinate of each point was uniformly randomly generated for the interval $[0, 1]$. We selected the set of dimensions $d=1, 2, 3, 4, 5, 10, 15,\ldots, 195, 200$ and the set of $l_p$ functionals $l_{0.01}, l_{0.1}, l_{0.5}, l_{1}, l_{2}, l_{4}, l_{10}, l_{\infty}$. For each dimension $d$ we prepared the database $X_d$ as the set of the first $d$ coordinates of points in database $X$. For each database $X_d$ and $l_p$ functional we calculate the set of all pairwise distances $D_{dp}$. Then we estimated the following values:

\begin{equation*}
\text{RC}_p=\frac{\max D_{dp} - \min D_{dp}}{\min D_{dp}}, \text{CV}_p=\frac{\sqrt{var(D_{dp})}}{mean(D_{dp})} .
\end{equation*}
Graphs of $\text{RC}_p$ and $\text{CV}_p$ are presented in Fig.~\ref{fig:RC1}. Fig.~\ref{fig:RC1} shows that our hypotheses are true. We can see that RC and CV as functions of dimension have qualitatively the same shape but in different scales: RC in the logarithmic scale.
The paper \cite{aggarwal2001} states that qualitatively different behaviour of $\max_i \|x_i\|_p-\max_i \|x_i\|_p$ for different $p$. We can state that for relative values we observe qualitatively the same  behaviour with small quantitative difference $\text{RC}_p-\text{RC}_q$ which decreases with dimensionality increasing. This means that that there could be some preference in usage of lower values of $p$ but the fractional metrics do not provide a panacea from the curse of dimensionality. To analyse this hypothesis, we study the real live benchmarks in the next section.

\section{Dimension estimation}\label{DimEst}
To consider high dimensional data and curse or blessing of dimensionality, it is necessary to determine what a dimension is. There are many different notions of dimension. Evaluation of dimension become very important with appearing of many ``big data'' databases. The number of attributes is dimension of vector space or Hamel dimension \cite{brown2012introduction} (denoted further as \#Attr). Fortunately, for the data mining problems the space dimension is not so important as a data dimension.
Intrinsic or effective dimension of data is not so well defined term by an obvious reason: the datasets are finite and, therefore, the direct application of the topological definitions of dimension gives zero. The most popular approach to definition of data dimension is approximation of data sets by a continuous topological object. Perhaps, the first and, at the same time, the widely used definition of intrinsic dimension is the dimension of linear manifold of ``the best fit to data'' with sufficiently small deviations \cite{pearson1901}.
The simplest way to calculate such dimension is Principle Component Analysis (PCA) \cite{jobson2012applied}. Unfortunately there is no unique methods to define number of informative (important, relevant, etc.) PCs \cite{jackson1993stopping}, \cite{ledesma2007determining}, \cite{cangelosi2007component}. Two most widely used methods are Kaiser rule \cite{guttman1954some} (denoted further as PCA-K), \cite{kaiser1960application} and Broken stick rule \cite{jackson1993stopping} (denoted further as PCA-BS).

Let us consider a database $X$ with $n$ data points $X={x_1,\ldots,x_n}$ and $d$ real-valued attributes, $x_i=(x_{i1},\ldots,x_{id})$.
Principal components correspond to eigenvalues of empirical covariance matrix $\Sigma(X)=X^TX$. This matrix is symmetric and non-negative defined. This means that eigenvalues of $\Sigma(X)$ matrix are non-negative real numbers. Let us denote these values as $\lambda_1\ge\lambda_2\ge\cdots\ge\lambda_d$. Fraction of Variance Explained (FVE) by $i$ PC is $f_i=\frac{\lambda_i}{\sum_{j=1}^{d}\lambda_j}$.

Kaiser rule states that all PC with FVE greater or equal to average FVE ($1/d$) are informative.

Let us consider unit interval (stick) randomly broken into $d$ fragments. Let us numerate this fragments with decreasing of their length: $s_1\ge s_2\ge\cdots\ge s_d$. Expected length of $i$ fragment is
\begin{equation}\label{eq.BS}
b_i=\frac{1}{d}\sum_{j=i}^{d}\frac{1}{j}.
\end{equation}
Broken stick rule states that the first $k$ PCs are informative, where $k$ is the maximal number such that $f_i\ge b_i, \forall i\le k$.

In many problems empirical covariance matrix degenerates. Let us consider projection of data onto the first $k$ PCs: $\hat{X}=XV$, where columns of matrix $V$ are the first $k$ eigenvectors of $\Sigma(X)$ matrix. Eigenvalues of the empirical covariance matrix $\Sigma(\hat(X))$ are $\lambda_1,\lambda_2,\ldots,\lambda_k$. After dimensionality reduction the condition number of the reduced covariance matrix should not be high in order to avoid the multicollinearity problems. The relevant definition \cite{gorban2018correction} of the intrinsic dimensionality refers directly to the conditional number of $\Sigma(X)$ matrix: $k$ is the number of the informative PCs if it is the smallest number such that
\begin{equation}\label{eq.CN}
\frac{\lambda_{k+1}}{\lambda_1}<\frac{1}{C},
\end{equation}
where C is specified conditional number, for example $C=10$.  This approach is referred  further as PCA-CN. The PCA-CN intrinsic dimensionality is defined as the number of eigenvalues of the covariance matrix that exceed a fixed percent of its largest eigenvalue  \cite{Fukunaga1971}.

Development of the idea of data approximation led to Principal Manifolds \cite{gorban2008principal} and more sophisticated approximators like Principal graphs and complexes \cite{gorban2007topological}, \cite{gorban2010principal}. These approaches provide tools for evaluating the intrinsic dimensionality of data and measuring the data complexity \cite{zinovyev2013data}. Another approach uses the complexes with vertices in data points: just connect the points with distance less than $\varepsilon$ for variable $\varepsilon$ and receive an object of combinatorial topology, a simplicial complex \cite{Carlsson2009topology}. All these methods use an object embedded into a dataspace. They are called \emph{Injective Methods} \cite{bac2020lizard}. Additionally, a family of Projective Methods was developed. They do not construct an approximator but project the dataspace on a lower dimensional space with preservation of objects similarity or dissimilarity. For a brief review of modern injective and projective methods we refer to \cite{bac2020lizard}.

Recent development of curse/blessing dimensionality studies introduce new method of intrinsic dimension evaluation separability analysis. Detailed description of this method can be found in \cite{gorban2018correction} and \cite{albergante2019estimating} (denoted further as SepD). For this study we used implementation of separability analysis from the \cite{matlabZin}. The main notion of this approach is $\alpha$ Fisher separability: point $x$ of dataset $X$ is $\alpha$ Fisher separable from dataset $X$ if
\begin{equation}\label{eq.Sep}
(x,y)\le\alpha (x,x), \forall y\in X, y\ne x,
\end{equation}
where $(x,y)$ is dot product of vectors $x$ and $y$.

The last used in this study intrinsic dimension is fractal dimension (denoted further as FracD). There are many versions of fractal dimension and we used R implementation from the RDimtools package \cite{RDimtools}. The considered definition of a fractal dimension is
$$d_f=\lim_{r\to 0}\frac{\log(N(r))}{\log(1/r)},$$
where $r$ is the $d$-cubic box size in the regular grid and $N(r)$ is number of cells with data points in this grid. Since we work with the finite dataset the limit is substituted by slope of linear regression without intercept.

\section{Comparison of $l_p$ functionals}\label{compar}

In Section~\ref{MesConc}, we demonstrated  that $\text{RC}_p$ is higher for smaller  $p$. Paper \cite{beyer1999} shows that greater RC means `more meaningful' task for kNN. We decide to compare different $l_p$ functions for kNN classification. Classification has one additional benefit in comparison with regression and clustering problems: classification quality measure is classifier independent and similarity measure independent \cite{sammut2011}.
For this study we selected three classification quality measures: Total Number of neighbours of the Same Class (TNNSC), accuracy (fraction of correctly recognised cases), sum of sensitivity (fraction of correctly solved cases of positive class) and specificity (fraction of correctly solved cases of negative class). TNNSC is not an obvious measure of classification quality and we use it for comparability of our results with \cite{aggarwal2001}. The 11 nearest neighbours as the method of classification was selected also for comparability with \cite{aggarwal2001}.

\subsection{Databases for comparison}

We selected 25 databases from UCI data repository \cite{Dua2017}. We applied the following criteria for the database selection:
\begin{enumerate}
  \item Data are not time-series.
  \item Database is formed for the binary classification problem.
  \item Database does not contain any missed values.
  \item Number of attributes is less than number of observations and is greater than 3.
  \item All predictors are binary or numeric.
\end{enumerate}

Totally, we selected 25 databases and 37 binary classification problems. For simplicity further we call each problem a `database'. The  list of selected databases is presented in Table~\ref{tab:DBs}.

\begin{table*}[tb]
\caption{Databases selected for analysis}
\begin{center}
\begin{tabular}{|l|c|r|r|r|r|r|r|r|}
\hline
\textbf{Name}& \textbf{Source}&\textbf{\#Attr.}&\textbf{Cases}&\textbf{PCA-K}&\textbf{PCA-BS}&\textbf{PCA-CN}&\textbf{SepD}&\textbf{FracD}\\
\hline
Blood & \cite{bloodDB} & 4 & 748 & 2 & 2 & 3 & 2.4 & 1.6\\ \hline
Banknote authentication & \cite{banknoteDB} & 4 & 1,372 & 2 & 2 & 3 & 2.6 & 1.9\\ \hline
Cryotherapy & \cite{khozeimeh2017}, \cite{khozeimeh2017a}, \cite{CryotherapyDB} & 6 & 90 & 3 & 0 & 6 & 4.1 & 2.5\\ \hline
Vertebral Column & \cite{VertebralDB} & 6 & 310 & 2 & 1 & 5 & 4.4 & 2.3\\ \hline
Immunotherapy & \cite{khozeimeh2017}, \cite{khozeimeh2017a}, \cite{ImmunotherapyDB} & 7 & 90 & 3 & 0 & 7 & 5.1 & 3.2\\ \hline
HTRU2 & \cite{HTRU2BD}, \cite{HTRU2BD2}, \cite{lyon2016} & 8 & 17,898 & 2 & 2 & 4 & 3.06 & 2.4\\ \hline
ILPD (Indian Liver &&&&&&&&\\
Patient Dataset) & \cite{ILDPDB} & 10 & 579 & 4 & 0 & 7 & 4.3 & 2.1\\ \hline
Planning Relax & \cite{RelaxDB} & 10 & 182 & 4 & 0 & 6 & 6.09 & 3.6\\ \hline
MAGIC Gamma Telescope & \cite{TelescopeDB} & 10 & 19,020 & 3 & 1 & 6 & 4.6 & 2.9\\ \hline
EEG Eye State & \cite{EEGDB} & 14 & 14,980 & 4 & 4 & 5 & 2.1 & 1.2\\ \hline
Climate Model Simulation &&&&&&&&\\
Crashes & \cite{ClimateDB} & 18 & 540 & 10 & 0 & 18 & 16.8 & 21.7\\ \hline
Diabetic Retinopathy Debrecen & \cite{DiabeticDB}, \cite{antal2014} & 19 & 1,151 & 5 & 3 & 8 & 4.3 & 2.3\\ \hline
SPECT Heart & \cite{HeartDB} & 22 & 267 & 7 & 3 & 12 & 4.9 & 11.5\\ \hline
Breast Cancer  & \cite{BreastDB} & 30 & 569 & 6 & 3 & 5 & 4.3 & 3.5\\ \hline
Ionosphere & \cite{IonosphereDB} & 34 & 351 & 8 & 4 & 9 & 3.9 & 3.5\\ \hline
QSAR biodegradation & \cite{mansouri2013}, \cite{QSARDB} & 41 & 1,055 & 11 & 6 & 15 & 5.4 & 3.1\\ \hline
SPECTF Heart & \cite{HeartDB} & 44 & 267 & 10 & 3 & 6 & 5.6 & 7\\ \hline
MiniBooNE particle &&&&&&&&\\
identification & \cite{MiniBooNEDB} & 50 & 130,064 & 4 & 1 & 1 & 0.5 & 2.7\\ \hline
First-order theorem proving &&&&&&&&\\
 (6 tasks) & \cite{bridge2014},  \cite{TheoremDB} & 51 & 6,118 & 13 & 7 & 9 & 3.4 & 2.04\\ \hline
Connectionist Bench (Sonar) & \cite{SonarDB} & 60 & 208 & 13 & 6 & 11 & 6.1 & 5.5\\ \hline
Quality Assessment of &&&&&&&&\\
Digital Colposcopies (7 tasks) & \cite{ColposcopiesDB},  \cite{fernandes2017} & 62 & 287 & 11 & 6 & 9 & 5.6 & 4.7\\ \hline
LFW & \cite{LFWTech} & 128 & 13,233 & 51 & 55 & 57 & 13.8 & 19.3\\ \hline
Musk 1 & \cite{MuskDB} & 166 & 476 & 23 & 9 & 7 & 4.1 & 4.4\\ \hline
Musk 2 & \cite{MuskDB} & 166 & 6,598 & 25 & 13 & 6 & 4.1 & 7.8\\ \hline
Madelon & \cite{guyon2005}, \cite{MadelonDB} & 500 & 2,600 & 224 & 0 & 362 & 436.3 & 13.5\\ \hline
Gisette & \cite{GisetteDB}, \cite{guyon2005} & 5,000 & 7,000 & 1465 & 133 & 25 & 10.2 & 2.04\\ \hline

\end{tabular}
\label{tab:DBs}
\end{center}
\end{table*}

We do not try to identify the best database preprocessing for each database. We simply use three preprocessing for each database:
\begin{itemize}
  \item empty preprocessing means usage data `as is';
  \item standardisation means to shift and scale data to have zero mean and unit variance;
  \item min-max normalization means to shift and scale data to belong interval $[0,1]$.
\end{itemize}

\subsection{Approaches to comparison}

Our purpose is to compare metrics but not to create the best classifier to solve each  problem. Following [1] we use 11NN  classifier. One of the reasons to select kNN is strong dependence of kNN on selected metrics and, on the other hand, the absence of any assumption about data, excluding the principle: tell me your neighbours, and will I tell you what you are. In our study we consider 11NN with $l_{0.01}, l_{0.1}, l_{0.5}, l_1, l_2, l_4, l_{10}, l_\infty$ as different algorithms.
We applied the following indicators to compare 11NN classifiers (algorithms) for listed $l_p$ functionals:
\begin{itemize}
  \item number of databases for which algorithm is the best \cite{demvsar2006statistical};
  \item number of databases for which algorithm is the worst \cite{demvsar2006statistical};
  \item number of databases for which algorithm has performance which statistically insignificantly different from the best;
  \item number of databases for which algorithm has performance which statistically insignificantly different from the worst;
  \item Friedman test \cite{friedman1940comparison}, \cite{friedman1937use} and post hoc Nomenyi test \cite{nemenyi1962distribution} which were specially developed for multiple algorithms comparison;
  \item Wilcoxon signed rank test was used for comparison of three pairs of metrics.
\end{itemize}

The first four approaches we call frequency comparison. To avoid discrepancies, a description of all statistical tests used is presented below.

\subsubsection{Proportion estimation}
Since two measures of classification quality accuracy and $\text{TNNSC}/(11\times n)$, where $n$ is number of cases in database, are proportions we can apply z-test of proportion estimations \cite{altman2013statistics}. We to compare two proportions with the same sample size, hence, we can use simplified formula for test statistics:
\begin{equation*}
z=\frac{|p_1 - p_2|}{\sqrt{\frac{p_1+p_2}{n}\big(1-\frac{p_1+p_2}{2}\big)}},
\end{equation*}
where $p_1$ and $p_2$ are two proportions to compare. \emph{P}-value of this test is probability of observing by chance the same or greater $z$ if both samples are taken from the same population. \emph{P}-value is $p_z=\Phi(-z)$, where $\Phi(z)$ is standard cumulative normal distribution.
We also meet the problem of reasonable selection of significance level. Selected databases contain from 90 to 130,064 cases. Usage of the same threshold for all databases is meaningless \cite{biau2008statistics}, \cite{kadam2010sample}. The necessary sample   size $n$ can be estimated through the specified significance level of $1-\alpha$, the statistical power $1-\beta$, the expected effect size $e$, and the population variance $s^2$. For the  normal distribution (since we use z-test):
\begin{equation*}
n=\frac{2(z_{1-\alpha}+z_{1-\beta})^2s^2}{d^2}.
\end{equation*}
In this study, we assume that the significance level is equal to the statistical power $\alpha=\beta$, the expected effect size is 1\% (1\% difference in accuracy is big enough), and the  population variance can be estimated by
\begin{equation*}
s^2=n\frac{n_+}{n}\big(1-\frac{n_+}{n}\big)=\frac{n_+(n-n_+)}{n},
\end{equation*}
where $n_+$ is number of cases in the positive class. Under this assumptions, we can estimate reasonable significance level as
\begin{equation*}
\alpha=\Phi\left(\frac{d}{s}\sqrt{\frac{n}{8}} \right).
\end{equation*}
Usage of 8 $l_p$ functionals means multiple testing. To avoid overdetection problem we apply Bonferroni correction \cite{weisstein2004bonferroni}. From the other side, usage of too big significance level is also meaningless \cite{biau2008statistics}. As a result we select the significance level as

\begin{equation*}
\alpha=\max \bigg \{\frac{1}{28}\Phi\bigg(\frac{d}{s}\sqrt{\frac{n}{8}} \bigg),0.00001\bigg \} .
\end{equation*}
Differences between two proportions (TNNSC or accuracies) is statistically significant if $p_z<\alpha$.
It is necessary to stress that for TNNSC the number of cases is $11n$ because we consider 11 neighbours for each point.

\subsubsection{Friedman test and post hoc Nomenyi test}
One of the widely used statistical tests for algorithms comparison on many databases is Friedman test \cite{friedman1940comparison}, \cite{friedman1937use}. To apply this test, we need firstly to apply tied ranking for the classification quality measure for one database: if several classifiers provide exactly the same quality measure then rank of all such classifiers will be equal to average value of the ranks for which they were tied \cite{friedman1937use}. Let us denote the number of used databases as $N$, the number of used classifiers as $m$ and the rank of classifier $i$ for database $j$ as $r_{ji}$. Mean rank of classifier $i$ is
\begin{equation*}
R_i=\frac{1}{N}\sum_{j=1}^{N} r_{ji}.
\end{equation*}
Test statistics is
\begin{equation*}
\chi^2_F=\frac{4N^2(m-1)\Big(\sum_{i=1}^{m} R_{i}^2 -\frac{m(m+1)^2}{4} \Big)}{4\sum_{i=1}^{m}\sum_{j=1}^{N} r_{ji}^2 -Nm(m+1)^2}.
\end{equation*}
Test statistics under null hypothesis that all classifiers have the same performance follows $\chi^2$ distribution with $m-1$ degrees of freedom. \emph{P}-value of this test is probability of observing by chance the same or greater $\chi^2_F$ if all classifiers have the same performance. \emph{P}-value is $p_\chi=1-F(\chi^2_F; m-1)$, where $F(\chi; df)$ is cumulative $\chi^2$ distribution with $df$ degrees of freedom. Since we have 37 databases only we decide to use 95\% significance level.

If Friedman test shows enough evidence to reject null hypothesis then we can conclude that not all classifiers have the same performance. To identify the pairs of classifiers with significantly different performance we applied post hoc Nomenyi test \cite{nemenyi1962distribution}. Test statistics for comparison of $i$ and $j$ classifiers is $|R_i-R_j|$. To identify pairs with statistically significant differences the critical distance
\begin{equation*}
\text{CD}=q_{\alpha m} \sqrt{\frac{m(m+1)}{6N}}.
\end{equation*}
is used. $q_{\alpha m}$ is critical value for Nomenyi test with significance level of $1-\alpha$ and $m$ degrees of freedom. The difference of classifiers performances is statistically significant with significance level of $1-\alpha$ if $|R_i-R_j|>CD$.

\subsubsection{Wilcoxon signed rank test}
To compare the performance of two classifiers on several databases we applied Wilcoxon signed rank test \cite{wilcoxon1945individual}. For this test we used standard Matlab function \textbf{signrank} \cite{matlabWilcoxon}.

\section{Dimension comparison}\label{DimComp}

We can see in Table~\ref{tab:DBs} that for all definitions of intrinsic dimension of data this dimension does not grow monotonically with the number of attributes for the given set of benchmarks. The correlation matrix of all six dimensions is presented in Table~\ref{tab:CorrDist}. There are two groups of highly correlated:
\begin{itemize}
\item \#Attr, PCA-K and PCA-BS;
\item PCA-CN and SepD.
\end{itemize}
The correlations between groups are low (maximal value 0.154). The last dimension - FracD - is correlated (but not strongly correlated) with PCA-CN and SepD.

\begin{table}[tb]
\caption{Correlation matrix for six dimensions: two groups of highly correlated estimates are highlighted by background}
\begin{center}
\scriptsize
\begin{tabular}{|l|r|r|r|r|r|r|r|}
\hline
\textbf{Dimension}&\textbf{\#Attr}&\textbf{PCA-K}&\textbf{PCA-BS}&\textbf{PCA-CN}&\textbf{SepD}&\textbf{FracD}\\\hline

\#Attr&\cellcolor{blue!30}{1.000}&\cellcolor{blue!30}{0.998}&\cellcolor{blue!30}{0.923}&0.098&0.065&-0.081\\ \hline
PCA-K&\cellcolor{blue!30}{0.998}&\cellcolor{blue!30}{1.000}&\cellcolor{blue!30}{0.917}&0.154&0.119&-0.057\\ \hline
PCA-BS&\cellcolor{blue!30}{0.923}&\cellcolor{blue!30}{0.917}&\cellcolor{blue!30}{1.000}&0.018&-0.058&0.075\\ \hline
PCA-CN&0.098&0.154&0.018&\cellcolor{green!30}{1.000}&\cellcolor{green!30}{0.992}&0.405\\ \hline
SepD&0.065&0.119&-0.058&\cellcolor{green!30}{0.992}&\cellcolor{green!30}{1.000}&0.343\\ \hline
FracD&-0.081&-0.057&0.075&0.405&0.343&1.000\\ \hline

\end{tabular}
\label{tab:CorrDist}
\end{center}
\end{table}

Let us consider the first group of correlated dimensions. Linear regressions PCA-K and PCA-BS of \#Attr are

\begin{equation*}\label{eq.LinRegK}
\begin{split}
\text{PCA-K} &= 0.29 \text{\#Attr},\\
\text{PCA-BS} &= 0.027 \text{\#Attr}.
\end{split}
\end{equation*}

It is necessary to emphasize that coefficient 0.29 (0.027 for PCA-BS) was defined for datasets considered in this study only and can be different for another datasets but multiple R squared equals 0.998 (0.855 for PCA-BS) shows that this dependence is not occasional. What is a reason of so strong correlations of these dimensions. It can be shown that these dimensions are sensitive to irrelevant or redundant features. The simplest example is adding of highly correlated attributes. To illustrate this property of considered dimensions let us consider abstract database $X$ with $d$ standardised attributes and covariance matrix $\Sigma$. This covariance matrix has $d$ eigenvalues $\lambda_1\ge\lambda_2\ge\ldots\ge\lambda_d$ and corresponding eigenvectors $v_1,\ldots,v_d$. To define PCA-K dimension we have to compare FVE of each PC with threshold $1/d$. Since all attributes are standardised we have unit values in the main diagonal of matrix $\Sigma$. This mean that $\sum_{i=1}^{d}\lambda_i=d$ and FVE of $i$ PC is $f_i=\frac{\lambda_i}{\sum_{j=1}^{d}\lambda_j}=\lambda_i/d$.

Consider duplication of attributes: add to the data table new attributes which are copies of the original attributes. This operation does not add any information to the data and, in principle, should not affect the internal dimension of the data for any reasonable definition.

Let us denote all object for this new database by $(1)$ in the superscript. New dataset can be denoted as $X^{(1)}=X|X$ where symbol $|$ denotes concatenation of two row vectors. For any data vectors $x^{(1)}$ and $y^{(1)}$ the dot product will be $(x^{(1)})^\intercal y^{(1)}=2x^\intercal y$, where $\intercal$ means matrix (vector) transpose.

For new dataset $X^{(1)}$ the covariance matrix has a form
\begin{equation*}\label{eq.Covar}
\Sigma ^{(1)}= \begin{bmatrix}
\Sigma & \Sigma\\
\Sigma & \Sigma
\end{bmatrix}.
\end{equation*}
The first $d$ eigenvectors can be presented in the form $v^{(1)}_i=(v_i^\intercal |v_i^\intercal)^\intercal$. Now we can calculate product of $v^{(1)}_i$ and $\Sigma ^{(1)}$:

\begin{equation*}\label{eq.Eigen}
\begin{split}
\Sigma ^{(1)}v^{(1)}_i= \begin{bmatrix}
\Sigma & \Sigma\\
\Sigma & \Sigma
\end{bmatrix}
\begin{pmatrix}
v_i \\
v_i
\end{pmatrix}&=
\begin{pmatrix}
\Sigma v_i + \Sigma v_i \\
\Sigma v_i + \Sigma v_i
\end{pmatrix}\\=
\begin{pmatrix}
\lambda_iv_i + \lambda_iv_i \\
\lambda_iv_i + \lambda_iv_i
\end{pmatrix}&=2\lambda_i
\begin{pmatrix}
v_i\\
v_i
\end{pmatrix}=
2\lambda_iv_i^{(1)}.
\end{split}
\end{equation*}
As we can see each of the first $d$ eigenvalues become twice greater ($\lambda^{(1)}_i=2\lambda_i, \forall i\le d$). This means that FVE of the first $d$ PCs have the same values
\begin{equation*}
f_i^{(1)}=\frac{\lambda^{(1)}_i}{2d}=\frac{2\lambda_i}{2d}=\frac{\lambda_i}{d}=f_i, \forall i\le d
\end{equation*}
Since sum of eigenvalues of matrix $\Sigma ^{(1)}$ is $2d$ we can conclude that $\lambda^{(1)}_i=0, \forall i>d$. We can repeat described procedure several times and define values $\lambda^{(m)}_i=m\lambda_i, f_i^{(m)}=f_i \forall i\le d$ and $\lambda^{(m)}_i=0, \forall i>d$, where $m$ is number of adding copies of attributes.
We need two more prepositions for broken stick. Let us have $d=2k$, then $b^{(1)}_{k+s}>b_{k+s}, \forall s>0, b^{(1)}_{k-s}<b_{k-s}, \forall s\ge 0$. Indeed, from \eqref{eq.BS}:

\begin{equation*}
\begin{split}
b^{(1)}_{k+s}&=\frac{1}{4k}\sum_{j=k+s}^{4k}\frac{1}{j}\\
&=\frac{1}{4k}\Bigg(\sum_{j=k+s}^{2k}\frac{1}{j}+\sum_{j=k+1}^{2k}\frac{1}{2j}+\sum_{j=k+1}^{2k}\frac{1}{2j-1}\Bigg)\\
&>\frac{1}{4k}\Bigg(\sum_{j=k+s}^{2k}\frac{1}{j}+\sum_{j=k+1}^{2k}\frac{1}{2j}+\sum_{j=k+1}^{2k}\frac{1}{2j}\Bigg)\\
&=\frac{1}{4k}\Bigg(\sum_{j=k+s}^{2k}\frac{1}{j}+\frac{1}{2}\sum_{j=k+1}^{2k}\frac{1}{j}+\frac{1}{2}\sum_{j=k+1}^{2k}\frac{1}{j}\Bigg)\\
&=\frac{1}{2}\Bigg(\frac{1}{2k}\sum_{j=k+s}^{2k}\frac{1}{j}+\frac{1}{2k}\sum_{j=k+1}^{2k}\frac{1}{j}\Bigg)=\frac{1}{2}(b_{k+s}+b_{k+1})\\
&\ge\frac{1}{2}(b_{k+s}+b_{k+s})=b_{k+s}
\end{split}
\end{equation*}
 and
\begin{equation*}
\begin{split}
b^{(1)}_{k-s}&=\frac{1}{4k}\sum_{j=k-s}^{4k}\frac{1}{j}\\
&=\frac{1}{4k}\Bigg(\sum_{j=k-s}^{2k}\frac{1}{j}+\sum_{j=k+1}^{2k}\frac{1}{2j}+\sum_{j=k+1}^{2k}\frac{1}{2j-1}\Bigg)\\
&<\frac{1}{4k}\Bigg(\sum_{j=k-s}^{2k}\frac{1}{j}+\sum_{j=k+1}^{2k}\frac{1}{2j}+\sum_{j=k+1}^{2k}\frac{1}{2j-2}\Bigg)\\
&=\frac{1}{4k}\Bigg(\sum_{j=k-s}^{2k}\frac{1}{j}+\sum_{j=k+1}^{2k}\frac{1}{2j}+\sum_{j=k+1}^{2k}\frac{1}{2j}-\frac{1}{4k}\Bigg)\\
&<\frac{1}{2}\Bigg(\frac{1}{2k}\sum_{j=k-s}^{2k}\frac{1}{j}+\frac{1}{2k}\sum_{j=k+1}^{2k}\frac{1}{j}\Bigg)\\
&=\frac{1}{2}(b_{k-s}+b_{k+1})
\le\frac{1}{2}(b_{k-s}+b_{k-s})=b_{k-s}
\end{split}
\end{equation*}
For case where $d=2k+1$ we have $b^{(1)}_{k+s}>b_{k+s}, \forall s>1, b^{(1)}_{k-s}<b_{k-s}, \forall s\ge -1$. Indeed, from \eqref{eq.BS}:
\begin{equation*}
\begin{split}
b^{(1)}_{k+s}&=\frac{1}{4k}\sum_{j=k+s}^{4k+2}\frac{1}{j}\\
&=\frac{1}{4k}\Bigg(\sum_{j=k+s}^{2k+1}\frac{1}{j}+\sum_{j=k+1}^{2k+1}\frac{1}{2j}+\sum_{j=k+2}^{2k+1}\frac{1}{2j-1}\Bigg)\\
&>\frac{1}{4k}\Bigg(\sum_{j=k+s}^{2k+1}\frac{1}{j}+\sum_{j=k+1}^{2k+1}\frac{1}{2j}+\sum_{j=k+2}^{2k+1}\frac{1}{2j}\Bigg)\\
&=\frac{1}{2}b_{k+s}+\frac{1}{4}b_{k+1}+\frac{1}{4}b_{k+2}\\
&>\frac{1}{2}b_{k+s}+\frac{1}{4}b_{k+s}+\frac{1}{4}b_{k+s}=b_{k+s}
\end{split}
\end{equation*}
 and
\begin{equation*}
\begin{split}
b^{(1)}_{k-s}&=\frac{1}{4k}\sum_{j=k-s}^{4k+2}\frac{1}{j}\\
&=\frac{1}{4k}\Bigg(\sum_{j=k-s}^{2k+1}\frac{1}{j}+\sum_{j=k+1}^{2k+1}\frac{1}{2j}+\sum_{j=k+2}^{2k+1}\frac{1}{2j-1}\Bigg)\\
&<\frac{1}{4k}\Bigg(\sum_{j=k-s}^{2k+1}\frac{1}{j}+\sum_{j=k+1}^{2k+1}\frac{1}{2j}+\sum_{j=k+2}^{2k+1}\frac{1}{2j-2}\Bigg)\\
&=\frac{1}{4k}\Bigg(\sum_{j=k-s}^{2k+1}\frac{1}{j}+\sum_{j=k+1}^{2k+1}\frac{1}{2j}+\sum_{j=k+1}^{2k}\frac{1}{2j}\Bigg)\\
&<\frac{1}{4k}\Bigg(\sum_{j=k-s}^{2k+1}\frac{1}{j}+\sum_{j=k+1}^{2k+1}\frac{1}{2j}+\sum_{j=k+1}^{2k+1}\frac{1}{2j}\Bigg)\\
&=\frac{1}{2}b_{k-s}+\frac{1}{2}b_{k+1}\le b_{k-s}
\end{split}
\end{equation*}

Now we are ready to evaluate the effect of duplication of attributes on the linear dimension estimate, bearing in mind that nothing should change for reasonable definitions.
\begin{itemize}
  \item For the vector space dimension we have simple formula: $\text{\#Attr}^{(m)}=md$.
  \item For Kaiser rule dimension PCA-K we have different threshold and now all PCs with FVE greater than $1/md$ are significant. This means that for all PCs with nonzero eigenvalues we can take big enough $m$ to provide ``informativeness'' of this PCs. Threshold of significance decays linearly with $m$ increasing.
  \item For broken stick dimension PCA-BS we observe initially increasing of threshold for the last half of original PCs but then thresholds $b^{(m)}_i$ will decay with $m$ increasing for all $i\le d$. This means that for all PCs with nonlinear eigenvalues we can take big enough $m$ to provide ``informativeness'' of this PCs. Thresholds of significance decays non-linearly with $m$ increasing. This slower than linear decay of thresholds provide less sensitivity to irrelevant attributes.
  \item For conditional number based dimension PCA-CN nothing changes with described procedure because of simultaneous multiplying of all eigenvalues by nonzero constant does not change fraction of eigenvalues in condition~\eqref{eq.CN}.
  \item For separability dimension adding of irrelevant features does not change anything because dot products of data points in the extended database are dot products of original dataset multiplied by $m$. This means that described extension of dataset change nothing in the separability~\eqref{eq.Sep}.
  \item There are no changes for the fractal dimension because of described extension of dataset do not change relative location of data points in space. This means that values $N(r)$ will be the same for original and extended datasets.
\end{itemize}

The second group of correlated dimensions includes PCA-CN, SepD, and FracD. The first two are extremely correlated and the last is moderately correlated with the first two. Linear regressions of these dimensions are

\begin{equation*}
\begin{split}
\text{SepD} &= 1.17 \text{PCA-CN},\\
\text{FracD} &= 0.052 \text{PCA-CN}.
\end{split}
\end{equation*}

High correlation of these three dimensions requires additional investigations.

\section{Results of $l_p$ functionals comparison}
Results of frequency comparison are presented in Table~\ref{tab:BestWorst}. Table~\ref{tab:BestWorst} shows that indicator `The best' is not robust and cannot be considered as a good tool for performance comparison \cite{demvsar2006statistical}. For example, for TNNSC with empty preprocessing $l_{0.1}$ is the best for 11 databases and it is maximal value but $l_{0.5}, l_1$ and $l_2$ are essentially better if we consider indicator `Insignificantly different from the best': 26 databases for $l_{0.1}$ and 31 databases for $l_{0.5}, l_1$ and $l_2$. Unfortunately we cannot estimate this indicator for `sensitivity plus specificity' quality measure by used way (it can be done by t-test). Analysis of Table~\ref{tab:BestWorst} shows that in average $l_{0.5}, l_1, l_2$ and $l_4$ are the best and $l_{0.01}$ and $l_\infty$ are the worst.
Results of Friedman and post hoc Nomenyi tests are presented in Table~\ref{tab:TestResults}. It can be seen that $l_{1}$ is the best for 6 of 9 tests and $l_{0.5}$ is the best for the remaining 3 tests. From the other side, performances of $l_{0.5}, l_1$ and $l_2$ are insignificantly different for all 9 tests.

\begin{table*}[tb]
\caption{Frequency comparison for TNNSC, accuracy and sensitivity plus specificity}
\begin{center}
\begin{tabular}{|l|r|r|r|r|r|r|r|r|}
\hline
\textbf{Indicator\textbackslash p for $l_p$ functional}&\textbf{0.01}&\textbf{0.1}&\textbf{0.5}&\textbf{1}&\textbf{2}&\textbf{4}&\textbf{10}&$\infty$\\
\hline
\multicolumn{9}{|c|}{\textbf{TNNSC}}\\ \hline
\multicolumn{9}{|c|}{Empty preprocessing}\\ \hline
The best & 2 & 11 & 5 & 10 & 7 & 1 & 1 & 1\\ \hline
The worst & 19 & 0 & 1 & 0 & 1 & 3 & 4 & 8\\ \hline
Insignificantly different from the best & 17 & 26 & 31 & 31 & 31 & 30 & 23 & 22\\ \hline
Insignificantly different from the worst & 34 & 23 & 17 & 19 & 21 & 21 & 25 & 29\\ \hline
\multicolumn{9}{|c|}{Standardisation}\\ \hline
The best & 0 & 5 & 10 & 11 & 6 & 2 & 1 & 1\\ \hline
The worst & 18 & 2 & 0 & 0 & 1 & 2 & 4 & 10\\ \hline
Insignificantly different from the best & 19 & 26 & 33 & 32 & 31 & 30 & 25 & 24\\ \hline
Insignificantly different from the worst & 35 & 24 & 20 & 19 & 20 & 21 & 25 & 28\\ \hline
\multicolumn{9}{|c|}{Min-max normalization}\\ \hline
The best & 1 & 5 & 10 & 13 & 4 & 6 & 1 & 3\\ \hline
The worst & 23 & 4 & 2 & 2 & 3 & 3 & 4 & 7\\ \hline
Insignificantly different from the best & 19 & 26 & 32 & 31 & 30 & 29 & 26 & 26\\ \hline
Insignificantly different from the worst & 36 & 24 & 22 & 21 & 22 & 22 & 26 & 26\\ \hline
\multicolumn{9}{|c|}{\textbf{Accuracy}}\\ \hline
\multicolumn{9}{|c|}{Empty preprocessing}\\ \hline
The best & 3 & 9 & 9 & 15 & 6 & 5 & 1 & 2\\ \hline
The worst & 13 & 3 & 1 & 2 & 4 & 4 & 9 & 14\\ \hline
Insignificantly different from the best & 29 & 31 & 34 & 35 & 35 & 35 & 33 & 30\\ \hline
Insignificantly different from the worst & 35 & 32 & 28 & 28 & 29 & 29 & 30 & 31\\ \hline
\multicolumn{9}{|c|}{Standardisation}\\ \hline
The best & 2 & 5 & 12 & 18 & 7 & 3 & 1 & 1\\ \hline
The worst & 13 & 4 & 0 & 0 & 2 & 6 & 7 & 13\\ \hline
Insignificantly different from the best & 30 & 31 & 34 & 34 & 33 & 31 & 32 & 30\\ \hline
Insignificantly different from the worst & 35 & 32 & 29 & 29 & 30 & 31 & 33 & 33\\ \hline
\multicolumn{9}{|c|}{Min-max normalization}\\ \hline
The best & 2 & 7 & 15 & 8 & 8 & 3 & 3 & 6\\ \hline
The worst & 18 & 6 & 3 & 4 & 5 & 9 & 8 & 8\\ \hline
Insignificantly different from the best & 30 & 31 & 34 & 33 & 33 & 32 & 31 & 32\\ \hline
Insignificantly different from the worst & 36 & 33 & 31 & 31 & 31 & 32 & 33 & 32\\ \hline

\multicolumn{9}{|c|}{\textbf{Sensitivity plus specificity}}\\ \hline
\multicolumn{9}{|c|}{Empty preprocessing}\\ \hline
The best & 4 & 8 & 7 & 12 & 7 & 5 & 1 & 1\\ \hline
The worst & 14 & 2 & 1 & 1 & 3 & 5 & 8 & 12\\ \hline
\multicolumn{9}{|c|}{Standardisation}\\ \hline
The best & 4 & 7 & 8 & 15 & 7 & 2 & 1 & 0\\ \hline
The worst & 13 & 3 & 0 & 0 & 2 & 5 & 4 & 15\\ \hline
\multicolumn{9}{|c|}{Min-max normalization}\\ \hline
The best & 5 & 8 & 13 & 6 & 9 & 3 & 4 & 5\\ \hline
The worst & 15 & 4 & 2 & 3 & 3 & 7 & 8 & 13\\ \hline

\end{tabular}
\label{tab:BestWorst}
\end{center}
\end{table*}

\begin{table*}[tb]
\caption{Results of the Friedman test and post hoc Nomenyi test}
\begin{center}
\begin{tabular}{|l|c|c|r|r|r|r|r|r|r|r|r|r|}
\hline
\textbf{Preprocessing}&\textbf{Quality}&\textbf{Friedman's}& \multicolumn{2}{|c|}{\textbf{The best $l_p$}}&\multicolumn{8}{|c|}{\textbf{Set of insignificantly different}}
\\ \cline{4-13}
&\textbf{measure}&\textbf{\emph{p}-value}&\multicolumn{1}{|c|}{\textbf{$p$}}&\multicolumn{1}{|c|}{\textbf{$R_i$}}&\textbf{0.01}&\textbf{0.1}&\textbf{0.5}&\textbf{1}&\textbf{2}&\textbf{4}&\textbf{10}&$\infty$\\
\hline

 & TNNSC & $<0.0001$ & 1 & 6.2639 &  & X & X & X & X & X &  & \\ \cline{2-13}
Empty & Accuracy & $<0.0001$ & 1 & 6.2639 &  & X & X & X & X &  &  & \\ \cline{2-13}
 & Se+Sp & $<0.0001$ & 0.5 & 6.0556 &  & X & X & X & X &  &  & \\ \hline
 & TNNSC & $<0.0001$ & 1 & 6.6944 &  &  & X & X & X &  &  & \\ \cline{2-13}
Standardisation & Accuracy & $<0.0001$ & 1 & 6.8056 &  &  & X & X & X &  &  & \\ \cline{2-13}
 & Se+Sp & $<0.0001$ & 1 & 6.4722 &  & X & X & X & X &  &  & \\ \hline
 & TNNSC & $<0.0001$ & 1 & 6.4722 &  &  & X & X & X & X &  & \\ \cline{2-13}
Min-max normalization & Accuracy & $<0.0001$ & 0.5 & 6.0000 &  & X & X & X & X &  &  & \\ \cline{2-13}
 & Se+Sp & $<0.0001$ & 0.5 & 6.0000 &  &  & X & X & X & X &  & \\ \hline

\end{tabular}
\label{tab:TestResults}
\end{center}
\end{table*}

\begin{table*}[tb]
\caption{\emph{P}-values of Wilcoxon test for different $l_p$ functions (left) and different type of preprocessing (right): E for empty, S for standardised and M for min-max normalization preprocessing, Se+Sp stands for sensitivity plus specificity}
\begin{center}
\begin{tabular}{|l|c|r|r|r| r|l|c|r|r|r|r|r|}
\cline{1-5} \cline{7-11}
\textbf{Preprocessing}&\textbf{Quality}&\multicolumn{3}{|c|}{\textbf{\emph{p}-value for $l_p$ and $l_q$}}&&\textbf{Quality} & \textbf{$p$ of $l_p$}&\multicolumn{3}{|c|}{\textbf{\emph{p}-value for pair of preprocessing}} \\ \cline{3-5} \cline{9-11}

&\textbf{measure}&\multicolumn{1}{|c|}{\textbf{0.5 \& 1}}&\multicolumn{1}{|c|}{\textbf{0.5 \& 2}}&\multicolumn{1}{|c|}{\textbf{1 \& 2}}& &\textbf{measure}&\multicolumn{1}{|c|}{\textbf{function}}&\multicolumn{1}{|c|}{\textbf{E \& S}}&\multicolumn{1}{|c|}{\textbf{E \& M}}&\multicolumn{1}{|c|}{\textbf{S \& M}} \\ \cline{1-5} \cline{7-11}

& TNNSC & 0.6348 & 0.3418 & 0.0469 & & & 0.5 & 0.5732 & 0.8382 & 0.6151 \\ \cline{2-5} \cline{8-11}
Empty & Accuracy & 0.9181 & 0.0657 & 0.0064 & & TNNSC & 1 & 0.9199 & 0.5283 & 0.1792 \\ \cline{2-5} \cline{8-11}
 & Se+Sp & 0.8517 & 0.0306 & 0.0022 & & & 2 & 0.9039 & 0.3832 & 0.1418\\ \cline{1-5} \cline{7-11}
& TNNSC & 0.3098 & 0.1275 & 0.0014 & & & 0.5 & 0.8446 & 0.5128 & 0.3217\\ \cline{2-5} \cline{8-11}
Standardised & Accuracy & 0.6680 & 0.0202 & 0.0017 & & Accuracy & 1 & 0.8788 & 0.0126 & 0.0091 \\ \cline{2-5} \cline{8-11}
 & Se+Sp & 0.8793 & 0.0064 & 0.0011 & & & 2 & 0.5327 & 0.3127 & 0.3436 \\ \cline{1-5} \cline{7-11}
Min-max & TNNSC & 0.7364 & 0.0350 & 0.0056 & & & 0.5 & 0.6165 & 0.2628 & 0.0644 \\ \cline{2-5} \cline{8-11}
normalization & Accuracy & 0.1525 & 0.0218 & 0.2002 & & Se+Sp & 1 & 0.5862 & 0.0054 & 0.0067 \\ \cline{2-5} \cline{8-11}
 & Se+Sp & 0.1169 & 0.0129 & 0.3042 & & & 2 & 0.6292 & 0.3341 & 0.4780 \\ \cline{1-5} \cline{7-11}

\end{tabular}
\label{tab:Wilcoxon}
\end{center}
\end{table*}

We compared 8 different $l_p$ functionals on 37 databases. Authors of \cite{aggarwal2001} formulated the  hypotheses that: (i) $l_1$ based kNN is better than $l_2$ based one and (ii) that the ``fractional'' metrics can further improve performance. We can test the differences between $l_{0.5}, l_1$ and $l_2$ based kNN by direct usage of Wilcoxon test. This comparison does not take into account the multiple testing. Results of comparisons are presented in Table~\ref{tab:Wilcoxon}. The left table shows that for most cases $l_{0.5}$ and $l_1$ based kNN have insignificantly different performances and for the most cases $l_2$ based kNN is slightly worse than the previous two. Right table shows, that $l_{0.5}$ and $l_2$ based kNN are insensitive to preprocessing (performances for both methods are not significantly different for different preprocessing). In contrast with these two methods, $l_1$ based kNN shows significant difference for min-max normalization preprocessing in comparison with two other preprocessing.

\section{Discussion}

In this paper, we tested the rather popular hypothesis that using the norms $l_p$ with $p<2$ (preferably $p=1$) or even the quasinorm $l_p$ with $1>p>0$ helps to overcome the curse of dimensionality.

Traditionally, the first choice of test datasets to analyse the curse or blessing of dimensionality is to use samples from some simple distributions: uniform distributions in balls, cubes, other convex compacts, or normal distributions (see for example, \cite{korn2001}, \cite{beyer1999}, \cite{hinneburg2000}, \cite{aggarwal2001}, \cite{aggarwal2001outlier}, \cite{radovanovic2010}, \cite{gorban2016}, etc.). Then the  generalsations are used like the product distributions in a cube (instead of the  uniform distributions) or log-concave distributions (instead of  normal distributions) \cite{Talagrand1995,GuedonMilman2011,gorban2018correction}. We used data sampled from the uniform distribution in the unit  cube for analysis of distribution of $l_p$ distances in high dimensions for various $p$.

Collections of 25   datasets from different sources (Table~\ref{tab:DBs}) was used for testing classifiers. The number of attributes (dimension of the data space) varied from 4 to 5,000.

For real-life datasets, the distributions are not just unknown -- there is doubt that the data are sampled from a more or less regular distribution. Moreover, we cannot always be sure that the concepts of probability distribution and statistical sampling are applicable. If we want to test any hypothesis about the curse or blessing of dimensionality and methods of working with high-dimensional data, then the first problem we face is this: What is data dimensionality? Beyond   regular distribution hypotheses, we cannot blindly assume that the data dimensionality is the same as the number of attributes. Therefore, the first task was to evaluate the internal dimension of all the data sets selected for testing.

We applied and compared five data dimensionality estimates:
\begin{itemize}
\item PCA with  Kaiser rule for determining the number of principal components to retain (PCA-K);
\item PCA with the broken stick rule for determining the number of principal components to retain (PCA-BS);
\item PCA with the condition number criterion for determining the number of principal components to retain (PCA-CN);
\item The Fisher separability dimension (SepD);
\item The fractal dimension (FracD).
\end{itemize}

We demonstrate  that both the Kaiser rule (PCA-K) and the broken stick rule (CA-BS) are very sensitive to adding duplicates of attributes. It can be easily shown that these dimensions are also very sensitive to adding of highly correlated attributes. In particular, for these estimates, the number of major principal components retained depend  on the `tail' of the minor components.

The conditional number criterion (PC-CN) gives much stabler results. The dimensionality estimates  based on the fundamental topological and geometric properties of the data set (the Fisher separability dimension, SepD,  and the fractal dimension, FracD) are less sensitive to adding of highly correlated attributes and insensitive to attribute duplicates.

PCA-K and PCA-BS estimates are highly correlated ($r>0.9$). Their correlations with the number of attributes are also very high (Table~\ref{tab:CorrDist}). Correlations of these estimates with three other estimates (PC-CN, SepD, and FracD) are much lower. PC-CN and SepD estimates are highly correlated ($r>0.9$), and their correlations with FracD are remarkable but not so high (see Table~\ref{tab:CorrDist}).

The dimensionality estimates  based on the fundamental topological and geometric properties of the data set (SepD and FracD) are less sensitive to adding of highly correlated attributes and insensitive to attribute duplicates.

The results of testing convinced us that   PC-CN and SepD estimates of the intrinsic data dimensionality are   more suitable for practical use than   PCA-K and PCA-BS estimates. The  FracD estimate may be also suitable. A detailed comparison with many other estimates is beyond the scope of this paper.

The selection of generally acceptable criteria is necessary to identify the benefits of using non-Euclidean norms and quasinorms $l_p$ ($2>p>0$). The Relative Contrast (RC) or the  Coefficient of Variation (CV) of high dimensional data are widely used. It was demonstrated on some examples (see, for example, \cite{hinneburg2000}, \cite{aggarwal2001}) that for $l_p$ norms or quasinorms the RC decreases with increasing dimension. It was also  shown \cite{aggarwal2001} that RC for $l_p$ functionals with lower $p$  is greater than for $l_p$ norms with greater $p$ (see Fig.~\ref{fig:RC1}.

Our tests for the datasets  sampled from a regular distribution (uniform distribution in a cube) confirm  this phenomenon. However Fig.~\ref{fig:RC1} shows that decreasing of $p$ cannot compensate (improve) the curse of dimensionality: the RC for high dimensional data and small $p$ will be less than for usual Euclidean distance in some smaller dimensional space. Behaviour of CV with dimension is similar to RC. Our experiments show that for both considered distance concentration measures inequalities $RC_p<RC_q, \forall p>q$ and $CV_p<CV_q, \forall p>q$ hold, where $p$ and $q$.

Authors of \cite{aggarwal2001} stated that \textit{ ``fractional distance metrics can significantly improve the effectiveness of standard clustering algorithms''}.

On the contrary, our tests on the collection of the benchmark datasets showed that there is no direct relation between distance concentration (e.g. RC or CV) and quality of classifiers: $l_{0.01}$ based kNN has one of the worst performance than $RC_p$ and $CV_p$ for greater $p$. Comparison of classification quality of 11NN classifiers for different $l_p$ functionals and for different databases shows that greater relative contrast does not mean higher quality.

Authors of \cite{aggarwal2001} found that $l_1$ \textit{``is consistently more preferable than the Euclidean distance metric for high dimensional data mining applications''}.
Our study partially confirmed the first finding: kNN with $l_1$ distance frequently demonstrates better performance in comparison with $l_{0.01}, l_{0.1}, l_{0.5}, l_{2}, l_{4}, l_{10}, l_{\infty}$ but this difference is not statistically significant.

Finally, performance of kNN classifiers on the basis of $l_{0.5}, l_1$ and $l_2$ functionals is statistically indistinguishable. Detailed pairwise comparison of $l_{0.5}, l_1$ and $l_2$ functionals shows that the performance of $l_1$ based kNN is more sensitive to used data preprocessing than $l_2$. There is no unique and unconditional leader in $l_p$ functionals for classification tasks. We can conclude, that $l_p$ based kNN with very small $p<0.1$ and very big $p>4$ are almost always worse than with $0.1<p<4$. Our wide test shows that for all used preprocessing and all considered classifier quality measures the  performance of $l_p$ based kNN classifiers for $l_{0.5}, l_1$ and $l_2$ are not statistically significantly different.

There remain many questions for further study: how the kNN classifier performance depends on the intrinsic data dimension? How can we measure this dimension? Can the number of $l_2$ based  major principal components be considered as a reasonable estimate of  the ``real'' data dimension or it is necessary to use $l_1$ based PCA? Recently developed PQSQ PCA \cite{gorban2018PQSQ} gives the possibility  to create PCA with various subquadratic functionals, including $l_p$ for $0<p\le 2$. The question about performance of clustering algorithms with different $l_p$ functionals remains still open.  This problem seems less clearly posed than for classification problems, because there are no unconditional criteria for ``correct clustering'' (or too many criteria that contradict each other), as expected for learning without supervision.

\bibliographystyle{IEEEtran}

\end{document}